\documentclass{article}
\usepackage{spconf,amsmath,graphicx}
\usepackage{hyperref}
\usepackage{adjustbox}
\usepackage{xcolor}
\usepackage{times}
\usepackage[utf8]{inputenc}
\usepackage[small]{caption}
\usepackage{booktabs}

\usepackage{amssymb}
\usepackage{makecell}
\usepackage{multicol}
\usepackage{multirow}
\definecolor{indiagreen}{rgb}{0.07, 0.53, 0.03}

\title{Evaluating COPY-BLEND Augmentation for Low Level Vision Tasks}
\name{Pranjay Shyam$^{\star}$ \qquad Sandeep Singh Sengar$^{\dagger}$ \qquad Kuk-Jin Yoon$^{\ddagger}$ \qquad Kyung-Soo Kim$^{\star}$}

\address{$^{\star}$ Mechatronics Systems and Control Lab, KAIST, Republic of Korea \\
         $^{\dagger}$ Department of Computer Science, University of Copenhagen, Denmark  \\
         $^{\ddagger}$ Visual Intelligence Lab, KAIST, Republic of Korea}
\begin{document}
%
\maketitle
\begin{abstract}
Region modification-based data augmentation techniques have shown to improve performance for high level vision tasks (object detection, semantic segmentation, image classification, etc.) by encouraging underlying algorithms to focus on multiple discriminative features. However, as these techniques destroy spatial relationship with neighboring regions, performance can be deteriorated when using them to train algorithms designed for low level vision tasks (low light image enhancement, image dehazing, deblurring, etc.) where textural consistency between recovered and its neighboring regions is important to ensure effective performance. In this paper, we examine the efficacy of a simple copy-blend data augmentation technique that copies patches from noisy images and blends onto a clean image and vice versa to ensure that an underlying algorithm localizes and recovers affected regions resulting in increased perceptual quality of a recovered image. To assess performance improvement, we perform extensive experiments alongside different region modification-based augmentation techniques and report observations such as improved performance, reduced requirement for training dataset, and early convergence across tasks such as low light image enhancement, image dehazing and image deblurring without any modification to baseline algorithm\footnote{This research was supported by KAIST-KU Joint Research Center, KAIST, Korea (N11200035). We gratefully acknowledge the GPU donation from NVidia used in this research. Codes for experiments conducted in this paper is available at \url{https://github.com/PS06/Copy_Blend}. Correspondences should be made at pranjayshyam@kaist.ac.kr}.

\end{abstract}
\section{Introduction}
\label{sec:intro}
\vspace{-1mm} 


\begin{figure}[!t]
\centering
\includegraphics[width=0.99\columnwidth, height=6cm]{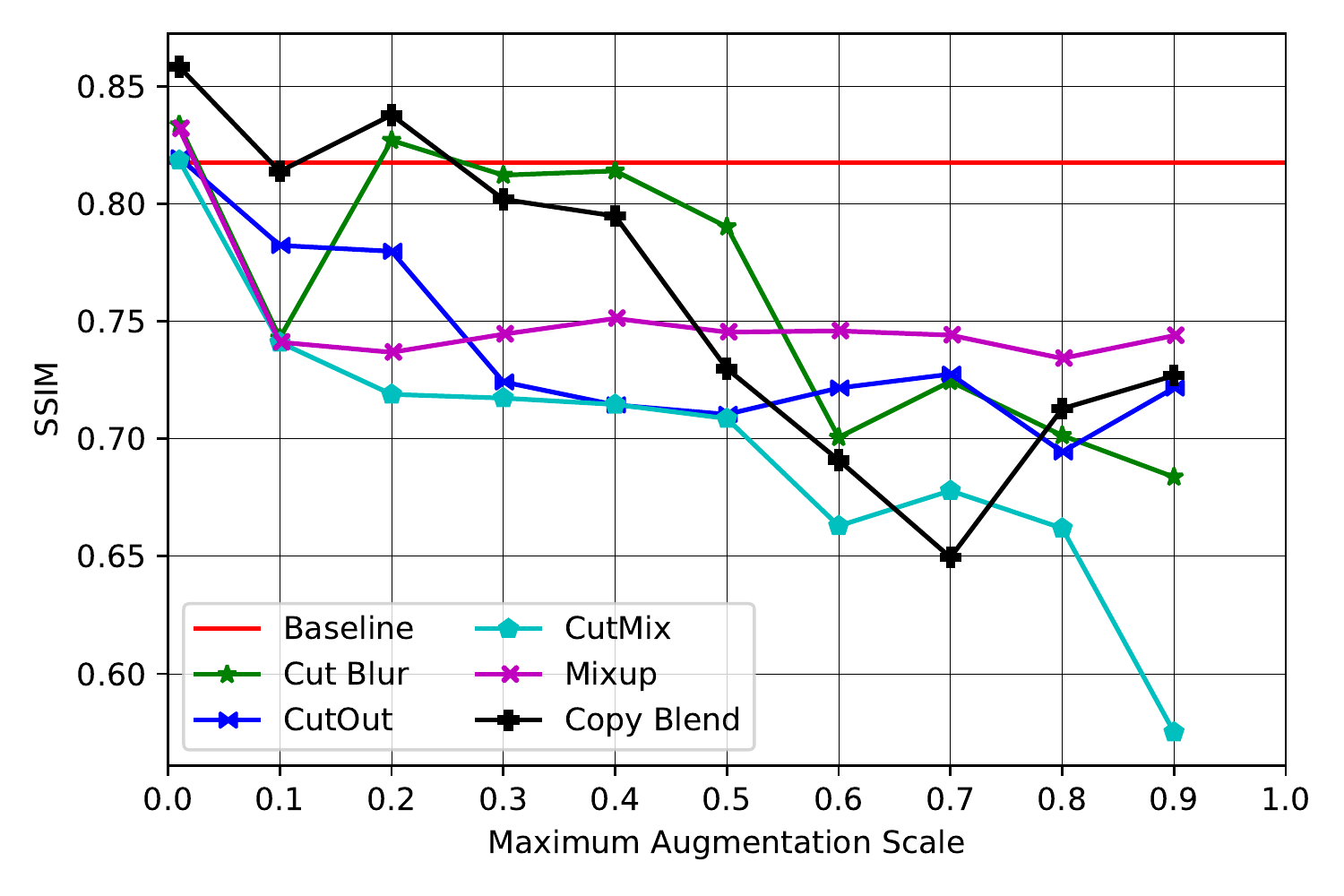}
\caption{Landscape (SSIM vs Maximum Augmentation Scale) demonstrating performance achieved by using different data augmentation strategies for Low Light Image Enhancement using DLN \cite{wang2020lightening} on LOL dataset \cite{Chen2018Retinex}.}
\vspace{-4mm}
\label{fig:fig_1}
\end{figure}

Data Augmentation (DA) is widely used as a regularization mechanism for ensuring robust and generalized performance of deep neural networks for high level tasks (e.g. image classification, object detection, instance segmentation) by increasing diversity within training set. Recent approaches \cite{singh2018hide, yun2019cutmix, zhang2017mixup, devries2017improved} focus on modification of certain regions within an image either by altering or dropping (setting pixel values to a fixed number) them to ensure underlying Convolutional Neural Network (CNN) is able to emphasize upon affected discriminative regions for generating desired predictions. However, this approach is not suited for low level tasks such as image enhancement or restoration (e.g. image dehazing, deblurring) wherein spatial consistency between recovered and neighboring pixels is necessary. Instead we observe that algorithms developed for such tasks when trained with augmentations distorting spatial properties result in poor performance (Fig. \ref{fig:fig_1}) compared to using simple augmentations like flipping and rotating (Baseline).

\begin{figure*}[!h]
\centering
\renewcommand{\tabcolsep}{1pt} 
\renewcommand{\arraystretch}{1} 
\begin{adjustbox}{width=0.99\linewidth}
\begin{tabular}{ccccccc}
\includegraphics[width=0.14\linewidth,height=2.5cm]{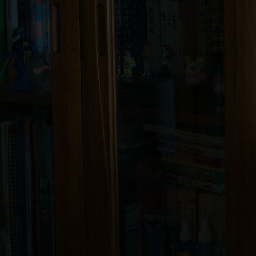} & 
\includegraphics[width=0.14\linewidth,height=2.5cm]{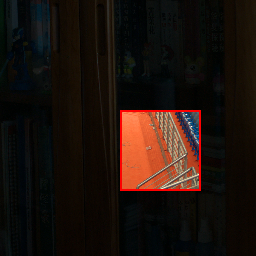} & 
\includegraphics[width=0.14\linewidth,height=2.5cm]{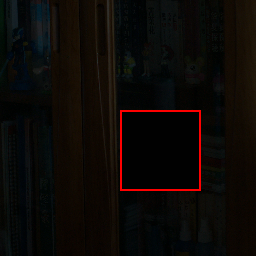} & 
\includegraphics[width=0.14\linewidth,height=2.5cm]{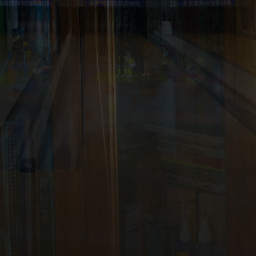} &
\includegraphics[width=0.14\linewidth,height=2.5cm]{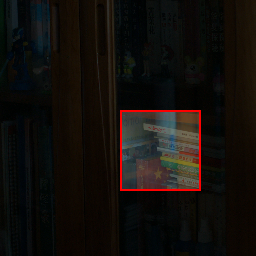} &
\includegraphics[width=0.14\linewidth,height=2.5cm]{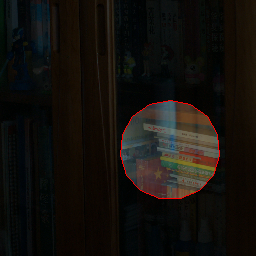} &
\includegraphics[width=0.14\linewidth,height=2.5cm]{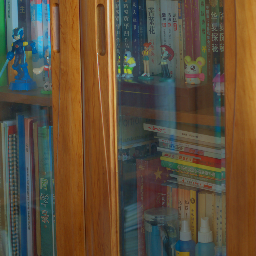} \\

Input & Cut Mix \cite{yun2019cutmix} & Cut Out \cite{devries2017improved} & Mixup \cite{zhang2017mixup} & Cut Blur \cite{yoo2020rethinking} & Copy Blend & GT \\

\end{tabular}
\end{adjustbox}
\vspace{-1mm}
\caption{Demonstration of different data augmentation techniques on the LOL dataset.}
\vspace{-1mm}
\label{fig:fig_2}
\end{figure*}

Furthermore enhancement/restoration tasks can exhibit diverse range of variations affecting different parts of an image, thus necessitating the underlying network be able to localize and determine the extent of enhancement or restoration necessary to generate a clean image. Ensuring such a distribution of variations within the training dataset is challenging and expensive, whereas using uniform distribution-based synthetic training sets does not ensure consistent performance in real conditions \cite{shyam2020domain}. Thus a DA technique that improves model performance (without increasing parameter count) by encouraging it to localize and determine extent of recovery without requiring additional data would be an ideal solution. Aiming to achieve such a mechanism for super resolution, Cut Blur\cite{yoo2020rethinking} proposed to cut patches from high resolution images, scale and paste onto a low/high resolution input image. However this approach results in sharp transitions and thus trains the model only for strong variations and avoids scenarios where weak intensity variations are present for events such as haze, motion blur, low light imaging. 
In this paper we extend this work and examine the efficacy of a simple copy-blend mechanism that copies patches of varying shapes and sizes from ground truth and blends onto noisy input at the same position and vice versa. This approach generates a diverse range of training samples with varying intensities of variation and thus ensures an underlying CNN emphasizes on where, what and how much the correction should be performed in order to closely match the recovered image with ground truth. We additionally observe that training CNNs for large epochs with copy-blend does not result in overfitting arising from data memorization, motivating us to explore if using the proposed mechanism allows for reducing the size of training set for enhancement and restoration tasks. 

\vspace{-2mm}
\section{Related Works}
\label{sec:related}
\vspace{-1mm}

Data augmentation techniques are gaining interest from research community primarily to reduce reliance on large scale well annotated datasets for training models with increased robustness and generalizability. An early approach, Mixup \cite{zhang2017mixup} mix labels and pixels from two samples within training set, essentially generating new synthetic samples containing labels of both samples. Cut-mix \cite{yun2019cutmix} extended mixup \cite{zhang2017mixup} concept by sampling and pasting rectangular patches and mixing labels proportional to patch area. While these techniques focus on modifying pixels, Cutout \cite{devries2017improved} proposed randomly setting pixels within a square patch to 0's resulting in an aggressive regularization effect. Though such methods improve model accuracy, it comes at the cost of model robustness. To alleviate such issues, Patch Gaussian \cite{lopes2019improving} proposed adding Gaussian noise patches randomly throughout an image. Motivated by the success of regional modification techniques, different works focused on leveraging them independently or combining a set of such methodologies to improve performance for other tasks such as object detection \cite{singh2018hide, zoph2019learning, dwibedi2017cut} and instance segmentation \cite{ghiasi2020simple}. 

To evaluate performance improvement achieved using these DA techniques, we consider tasks such as image dehazing, deblurring and low light image enhancement (LLIE). For our evaluation purpose, we use \{MSNet \cite{msnet2020}, DIDH \cite{shyam2020domain}\}, \{AFNet \cite{shyam2020adver}, DLN \cite{wang2020lightening}\}, \{DeblurGANv2 \cite{Kupyn_2019_ICCV}, DMPHN \cite{Zhang_2019_CVPR}\} algorithms for the task of dehazing, deblurring and LLIE as they represent SoTA algorithms and use their original implementation.

\vspace{-2mm}
\section{Copy Blend Algorithm}
\label{sec:problem}
\vspace{-1mm}

Given a data pair $(I_{IN}, I_{OUT})$ depicting noisy image and its corresponding clean image, a floating point mask $(\alpha \in [0, 1])$ representing regions and intensities to be copied from a clean image onto a noisy input image is constructed. The new input image $(I'_{IN})$ is constructed by blending two images following the relation, 

\vspace{-2mm}
\begin{equation}
    I'_{IN} = (1-\alpha)*I_{IN} + \alpha*I_{OUT}
\end{equation}

\begin{figure*}[!ht]
\centering
\renewcommand{\tabcolsep}{1pt} 
\renewcommand{\arraystretch}{1} 
\begin{adjustbox}{width=0.99\linewidth}
\begin{tabular}{cccc}

\includegraphics[width=0.25\linewidth, height=2.5cm]{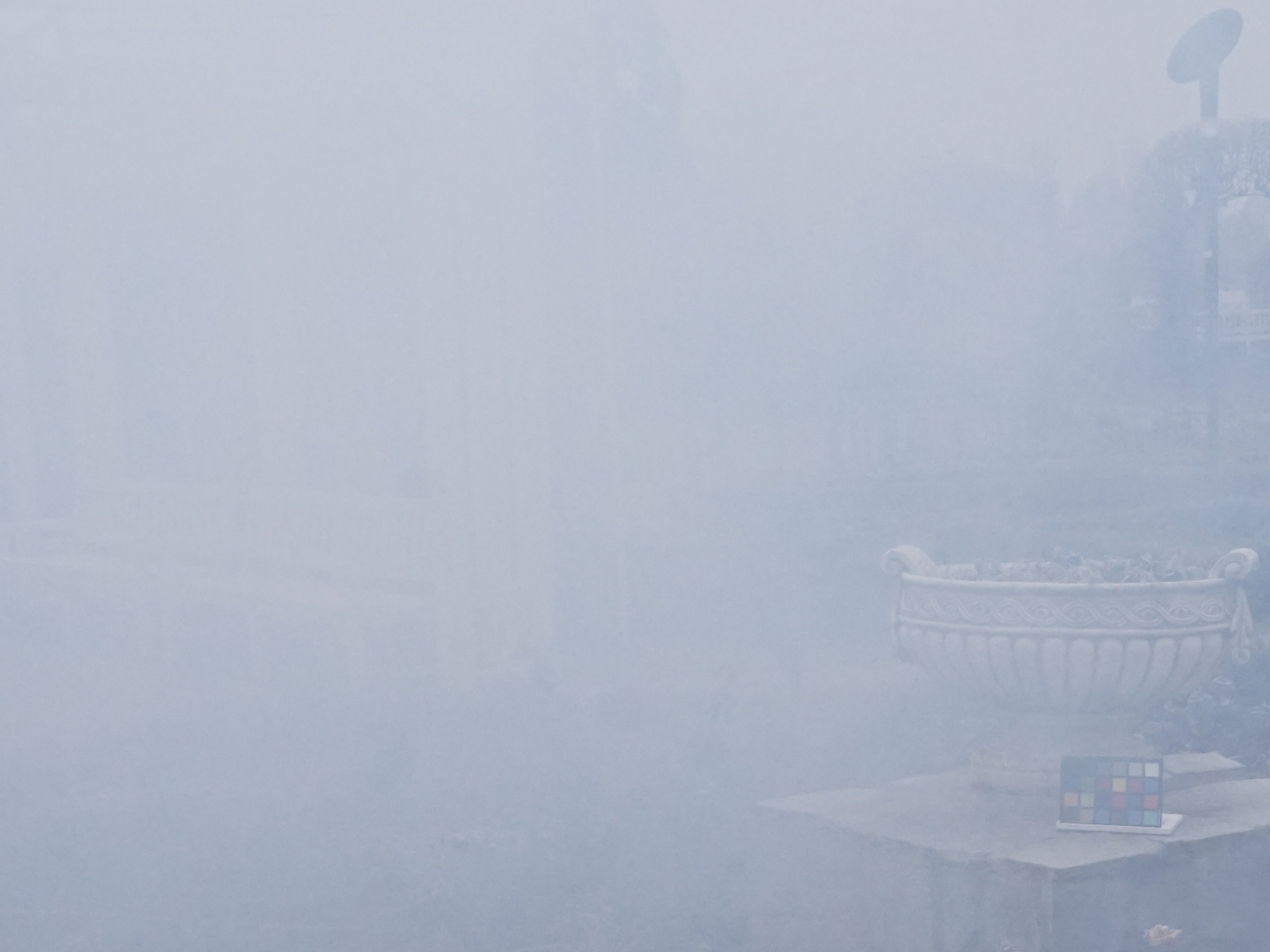} &
\includegraphics[width=0.25\linewidth, height=2.5cm]{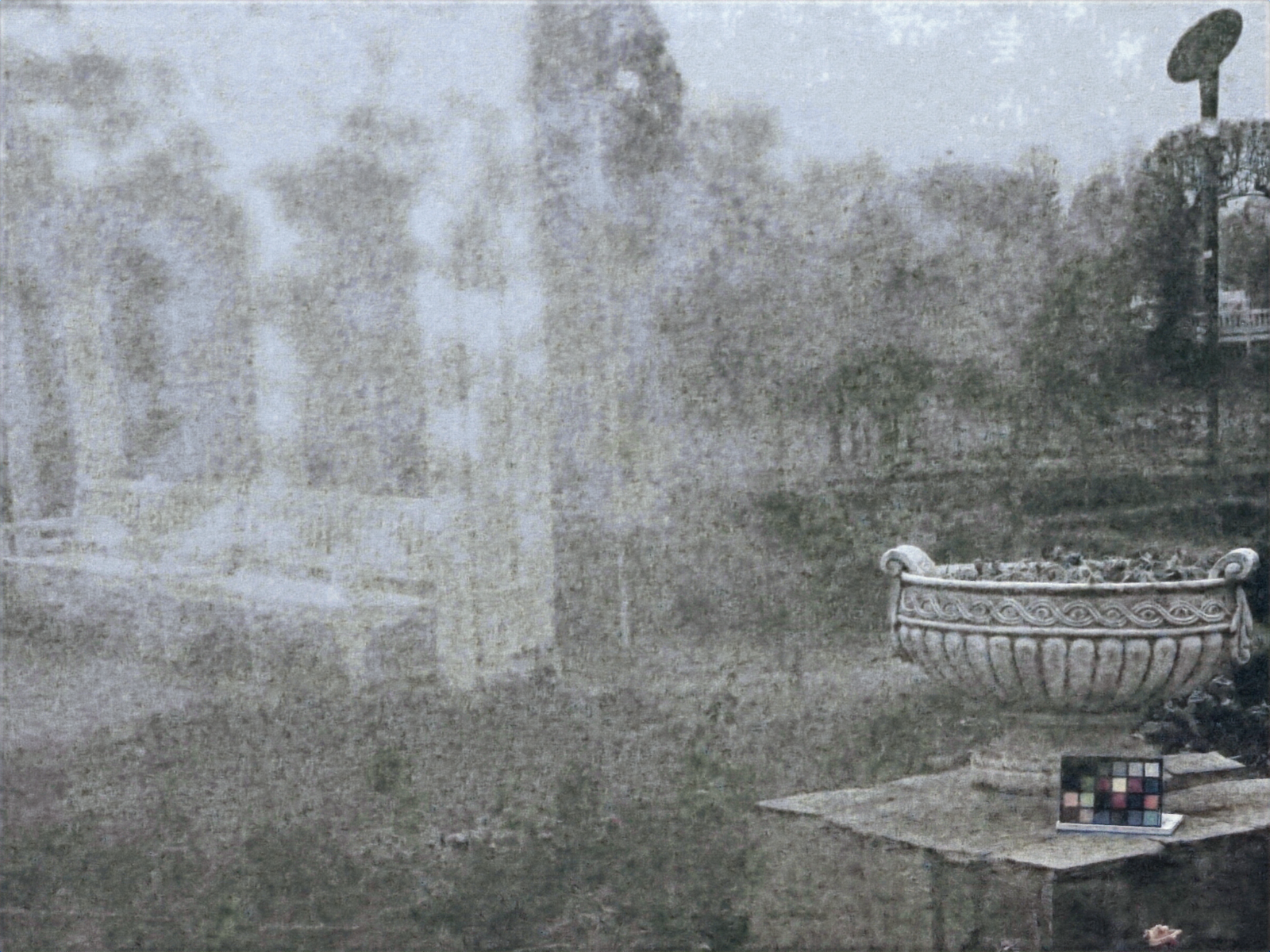} &
\includegraphics[width=0.25\linewidth, height=2.5cm]{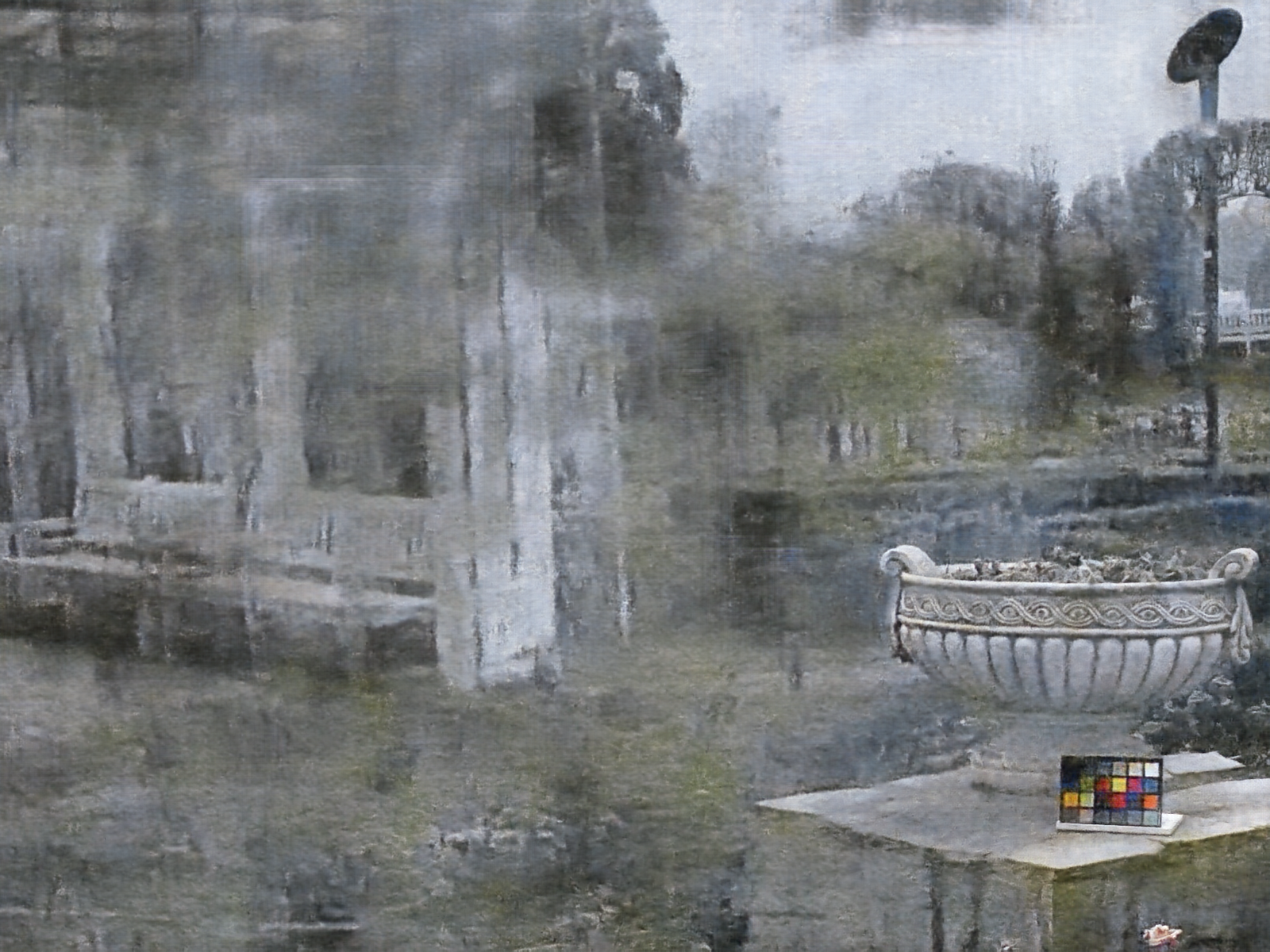} &
\includegraphics[width=0.25\linewidth, height=2.5cm]{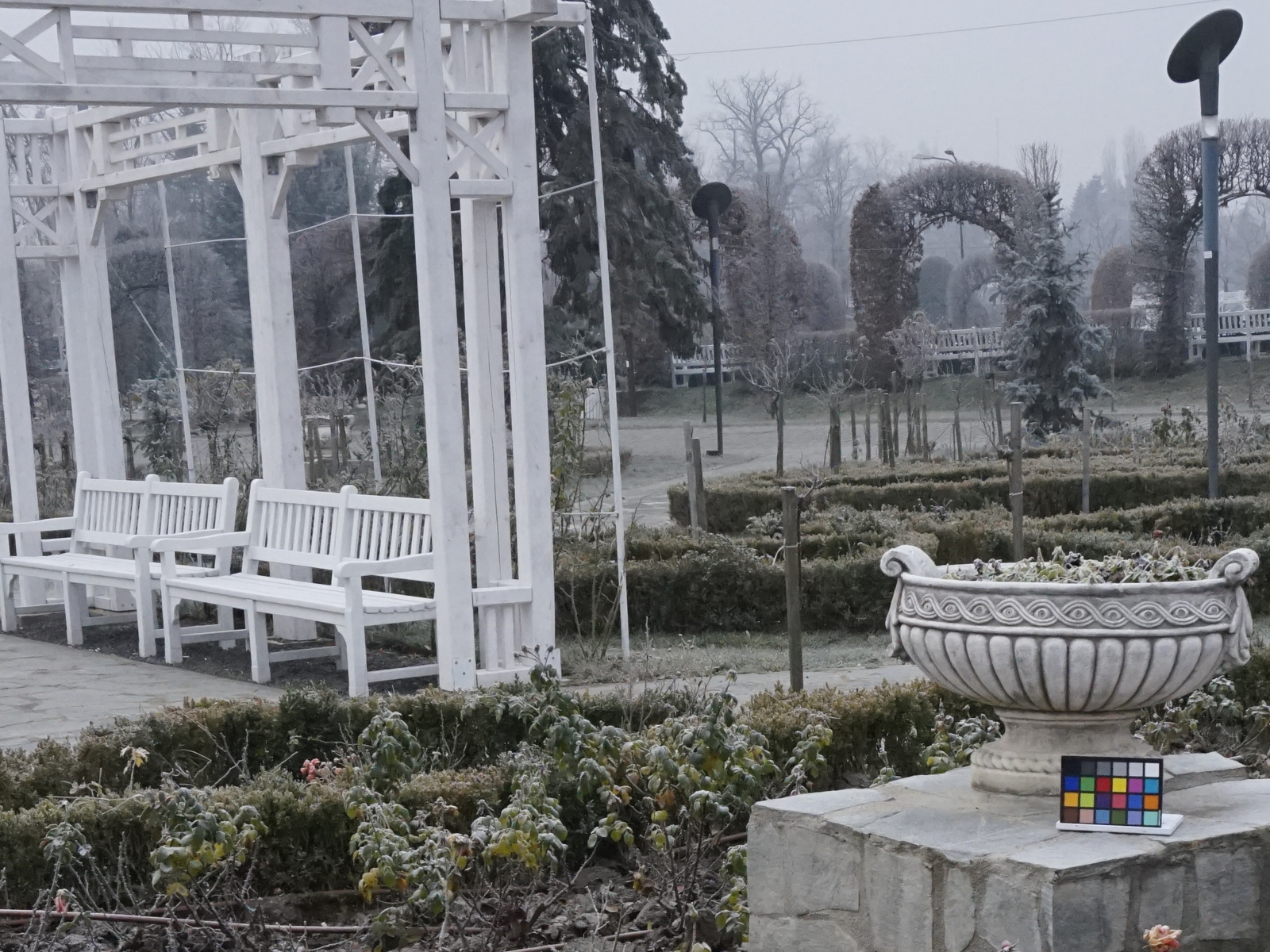} \\
9.75 / 0.41 / 5.45 & 15.04 / 0.42 / 4.17 & 15.59 / 0.47 / 3.54 & - / - / 1.85 \\
Input Hazy Image & MSNet & MSNet + CB & Output Image \\

\includegraphics[width=0.25\linewidth, height=2.5cm]{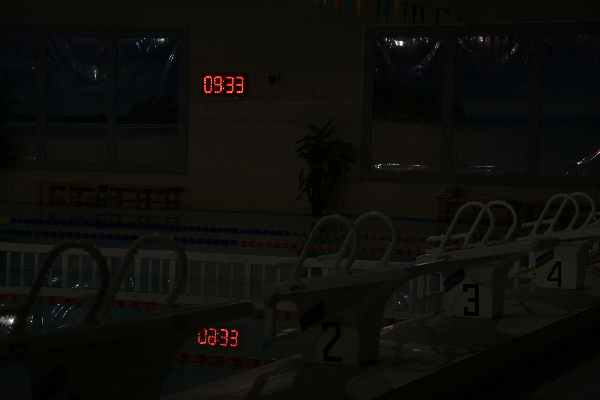} & 
\includegraphics[width=0.25\linewidth, height=2.5cm]{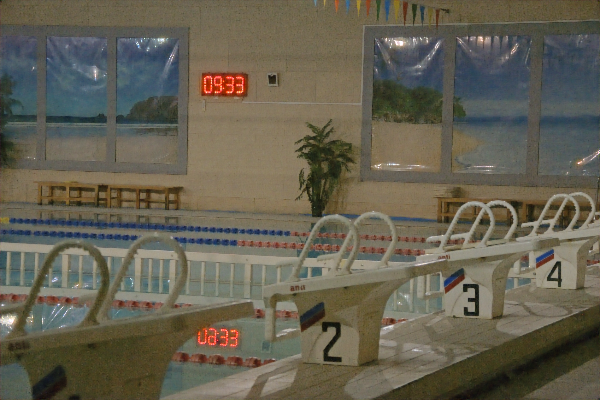} &
\includegraphics[width=0.25\linewidth, height=2.5cm]{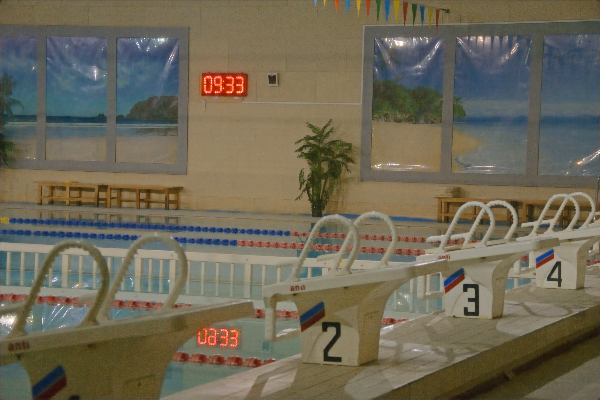} &
\includegraphics[width=0.25\linewidth, height=2.5cm]{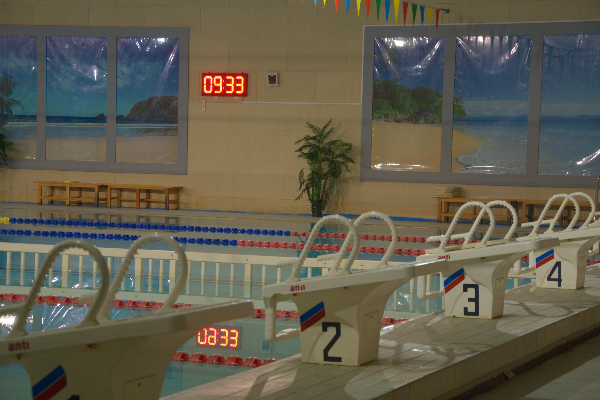} \\ 
9.96 / 0.18 / 5.35 & 28.67 / 0.86 / 3.48 & 26.04 / 0.89 / 3.62 & - / - / 4.09 \\
Input Low Light Image & DLN & DLN + CB & Output Image \\


\end{tabular}
\end{adjustbox} \vspace{-4pt}
\caption{Enhancement and restoration results of different algorithms trained within strong baseline formulation with and without the copy blend augmentation and corresponding performance metrics (PSNR / SSIM / NIQE). We can observe increased perceptual and structural quality in enhanced images.}
\vspace{-3mm}
\label{fig:fig_3}
\end{figure*}

In the proposed mechanism, we construct mask using two control parameters namely maximum patch size $(MP_{MAX})$ and blend intensity $(\beta \in [0, 1])$. This allows us to introduce patches of random numbers and sizes and distribute it randomly across image, generating non-homogeneous distribution. Since blending is performed on same location between input and output images, we can set  maximum patch size to match spatial resolution of an image, however we observe from Fig. \ref{fig:fig_1} that keeping maximum augmentation higher than a threshold doesn't aid in improving performance. Furthermore, we observe (Sec. 4.4) that varying patch shapes and number of patches does not improve performance substantially. Hence for our experiments, we set the number of square patches as 1 with maximum patch size of 0.2 w.r.t training image.

\vspace{-2mm}
\section{Experimental Evaluation}
\label{sec:experiments}
\vspace{-1mm}
\subsection{Datasets and Evaluation Metrics}
\vspace{-1mm}

In order to evaluate the effect of different algorithms for dehazing, deblurring, and LLIE, we utilize twin datasets to establish performance consistency and utilize NTIRE-19 \cite{cai2019ntire} and NTIRE-20 \cite{yuan2020ntire} datasets for dehazing, GO PRO \cite{Nah_2017_CVPR} and Real Blur \cite{rim_2020_ECCV} datasets for deblurring, and LOL \cite{Chen2018Retinex} and SICE \cite{Cai2018deep} datasets for LLIE. Subsequently we utilize traditional pixel and structural metrics such as Peak-Signal-to-Noise-Ratio (PSNR, capturing relationship between two images at pixel level on log scale), Structural Similarity Index Measure (SSIM, capturing structural similarity using luminance, contrast and structure between two images) and Image Quality Metrics such as Naturalness Image Quality Evaluator (NIQE, no reference metric to determine naturalness of an image) metrics and perform experiments on a computer equipped with Intel 8700K CPU, Titan V GPU and Pytorch 1.6 without any modifications to training settings of evaluation algorithms. For a comprehensive overview of different datasets, we summarize their properties in Table \ref{tab:tab_1}. In order to construct a strong baseline, we mix synthetic and real samples to train an underlying algorithm that is shown to improve algorithm performance and has been studied for dehazing in \cite{shyam2020domain} and LLIE in \cite{wang2020lightening}.

\vspace{-3mm}
\begin{table}[t]
    \caption{Properties of different datasets} \vspace{-5pt}
    \begin{adjustbox}{max width=0.99\columnwidth}
    \begin{tabular}{lccc}
    \Xhline{2\arrayrulewidth} \hline \noalign{\vskip 1pt}
    Dataset Name &  PSNR / SSIM / NIQE & Resolution & \# Val Images \\ 
    \Xhline{2\arrayrulewidth} \hline \noalign{\vskip 1pt}
    NTIRE-19  &  9.11 / 0.49 / 5.04 & 1600 $\times$ 1200 & 10 \\
    NTIRE-20  & 10.42 / 0.46 / 2.24 & 1600 $\times$ 1200 & 5 \\
    GO PRO    & 25.64 / 0.79 / 2.70 & 1280 $\times$  720 & 1111 \\
    Real Blur & 26.55 / 0.80 / 4.97 &  675 $\times$  769 & 980\\
    LOL       &  7.77 / 0.19 / 5.71 &  600 $\times$  400 & 15 \\
    SICE      & 12.26 / 0.57 / 4.13 & 5472 $\times$ 3648 & 58 \\
    \Xhline{2\arrayrulewidth} \hline
    \end{tabular}
    \end{adjustbox}
    \vspace{-3mm}
    \label{tab:tab_1}
\end{table}

\vspace{-1mm}
\subsection{Effect of Various Augmentation Techniques}
\vspace{-1mm}

In order to identify best suited augmentation scale for various techniques, we examine the relationship between an augmentation scale and performance on LLIE using LOL dataset and DLN algorithm, summarizing results in Fig. \ref{fig:fig_1}. For representing LLIE results we use feature metrics because pixel metrics (such as PSNR) assume pixel independent relationship and hence are poor metrics for measuring structural properties of an image. We observe that max augmentation scale of 0.2 results in the peak performance for nearly all DA techniques and hence we fix its value at 0.2 for subsequent experiments. We then extend the examination towards other tasks (dehazing and deblurring) and evaluate performance using MSNet and DeblurGANv2 algorithms while summarizing qualitative results in Table \ref{tab:tab_2}. When applying cut-out natively, we observe performance of LLIE algorithms to improve, however for dehazing and deblurring the performance reduces drastically. We subsequently changed the pixel value of cut-out regions to 128 (gray pixels), representing haze conditions and observed dehazing performance to improve.

\begin{table}[t]
\caption{Peak performance achieved by DLN, MSNet and DeblurGANv2 under different augmentation techniques} \vspace{-5pt}
\begin{adjustbox}{width=0.97\linewidth}
\begin{tabular}{lccc}
\Xhline{2\arrayrulewidth} \hline \noalign{\vskip 1pt}
\multirow{2}{5em}{Algorithm} & DLN & MSNet & DeblurGANv2 \\
\cline{2-4} \noalign{\vskip 1pt}
 & PSNR / SSIM & PSNR / SSIM  & PSNR / SSIM \\
\Xhline{2\arrayrulewidth} \hline \noalign{\vskip 1pt}
Baseline & 21.33 / 0.81 & 13.32 / 0.53 & 29.55 / 0.93 \\
CutMix  & \textcolor{red}{20.78} / \textcolor{indiagreen}{0.83} & \textcolor{indiagreen}{13.51 / 0.54} & \textcolor{red}{29.17 / 0.91} \\
Mixup & \textcolor{red}{20.57} / 0.81 & \textcolor{red}{13.05 / 0.49} & \textcolor{red}{29.23 / 0.91} \\
Cut Blur & \textcolor{indiagreen}{21.39 / 0.83} & \textcolor{indiagreen}{13.77 / 0.60} & \textcolor{indiagreen}{29.99 / 0.94} \\
Cut Out & \textcolor{indiagreen}{21.42} / 0.81 & \textcolor{indiagreen}{13.75 / 0.58} & \textcolor{red}{29.51 / 0.92} \\
Copy Blend & \textcolor{indiagreen}{21.47 / 0.86} & \textcolor{indiagreen}{14.27 / 0.62} & \textcolor{indiagreen}{29.91} / 0.93 \\
\Xhline{2\arrayrulewidth} \hline
\end{tabular}
\end{adjustbox}
\vspace{-4mm}
\label{tab:tab_2}
\end{table}

Hence we corroborate that cut-out techniques can be modified according to task, for improving model performance. As for extreme augmentations such as cut-mix and mix-up, we observe reduction in performance owing to breaking spatial characteristics between different regions. Furthermore, copy-blend and cut-blur augmentations tend to improve performance, with copy-blend improving the performance substantially specifically for dehazing and deblurring tasks, which we believe to arise from the ability to ensure varying intensity of degradations unlike copy-blur that results in strong degradations. We present some visual results in Fig. \ref{fig:fig_3} and elaborated quantitative results in Tab. \ref{tab:tab_3}. From these observations, we conclude that low level vision related tasks require spatial consistency within an input image for efficient training. One noticeable exception is of cut-out that improves model performance on LLIE and dehazing tasks but deteriorates model performance on deblurring. We presume this to be the outcome of cut regions representing extremely dark/hazy regions and hence aiding in improving performance, however the cut regions for deblurring represent lost information and thus deteriorate the performance.

\begin{table}[t]
\caption{Extended Evaluation of Copy Blend Augmentation} \vspace{-5pt} 
\begin{adjustbox}{width=0.97\linewidth}
\begin{tabular}{lcc}
\Xhline{2\arrayrulewidth} \hline \noalign{\vskip 1pt}
\multirow{2}{5em}{Algorithm} & PSNR / SSIM / NIQE & PSNR / SSIM / NIQE \\
\cline{2-3} \noalign{\vskip 1pt}
 & LOL & SICE \\
\Xhline{2\arrayrulewidth} \hline \noalign{\vskip 1pt}
DLN \cite{wang2020lightening} & 21.34 / 0.82 / 3.05 & 16.44 / 0.60 / 2.32 \\
+ CB                          & \textcolor{indiagreen}{21.47 / 0.84 / 2.84} & \textcolor{indiagreen}{ 16.51 / 0.62} / \textcolor{red}{2.37} \\
AFNet \cite{shyam2020adver}   & 20.17 / 0.81 / 3.17 & 18.75 / 0.64 / 2.42 \\
+ CB                          & \textcolor{indiagreen}{ 20.84 / 0.84 / 2.73 } & \textcolor{indiagreen}{ 18.91 / 0.65} / \textcolor{red}{2.49}\\
\Xhline{\arrayrulewidth} \hline \noalign{\vskip 1pt}
 & NTIRE-19 & NTIRE-20 \\
\Xhline{\arrayrulewidth} \hline \noalign{\vskip 1pt}
MSNet \cite{msnet2020}      & 13.32 / 0.53 / 4.21 & 12.04 / 0.50 / 4.08 \\
+ CB               & \textcolor{indiagreen}{ 14.71 / 0.58 / 3.87} & \textcolor{indiagreen}{13.97 / 0.57 / 3.77}\\
DIDH \cite{shyam2020domain}     & 15.71 / 0.54 / 4.71 & 14.71 / 0.45 / 5.34 \\
+ CB                    & \textcolor{indiagreen}{ 17.18 / 0.62 / 3.47} & \textcolor{indiagreen}{ 18.16 / 0.69 / 3.28 } \\
\Xhline{\arrayrulewidth} \hline \noalign{\vskip 1pt}
 & GO-PRO & Real Blur \\
\Xhline{2\arrayrulewidth} \hline \noalign{\vskip 1pt}
DeblurGANv2 \cite{Kupyn_2019_ICCV} & 29.55 / 0.93 / 3.13 & 28.70 / 0.86 / 3.49\\
+ CB                       & \textcolor{indiagreen}{ 29.91} / 0.93 / \textcolor{indiagreen}{3.07} & \textcolor{indiagreen}{ 31.26 / 0.92 / 3.19 } \\
DMPHN \cite{dong2020deep}  & 30.21 / 0.93 / 2.64 &  29.71 / 0.93 / 2.76 \\
+ CB                       & 30.21 / \textcolor{indiagreen}{0.94} / \textcolor{red}{2.73} & \textcolor{indiagreen}{ 31.18 / 0.94 / 2.50 } \\
\Xhline{2\arrayrulewidth} \hline
\end{tabular}
\end{adjustbox}
\vspace{-2mm}
\label{tab:tab_3}
\end{table}

\begin{figure}[t]
\centering
\includegraphics[width=0.99\columnwidth, height=6cm]{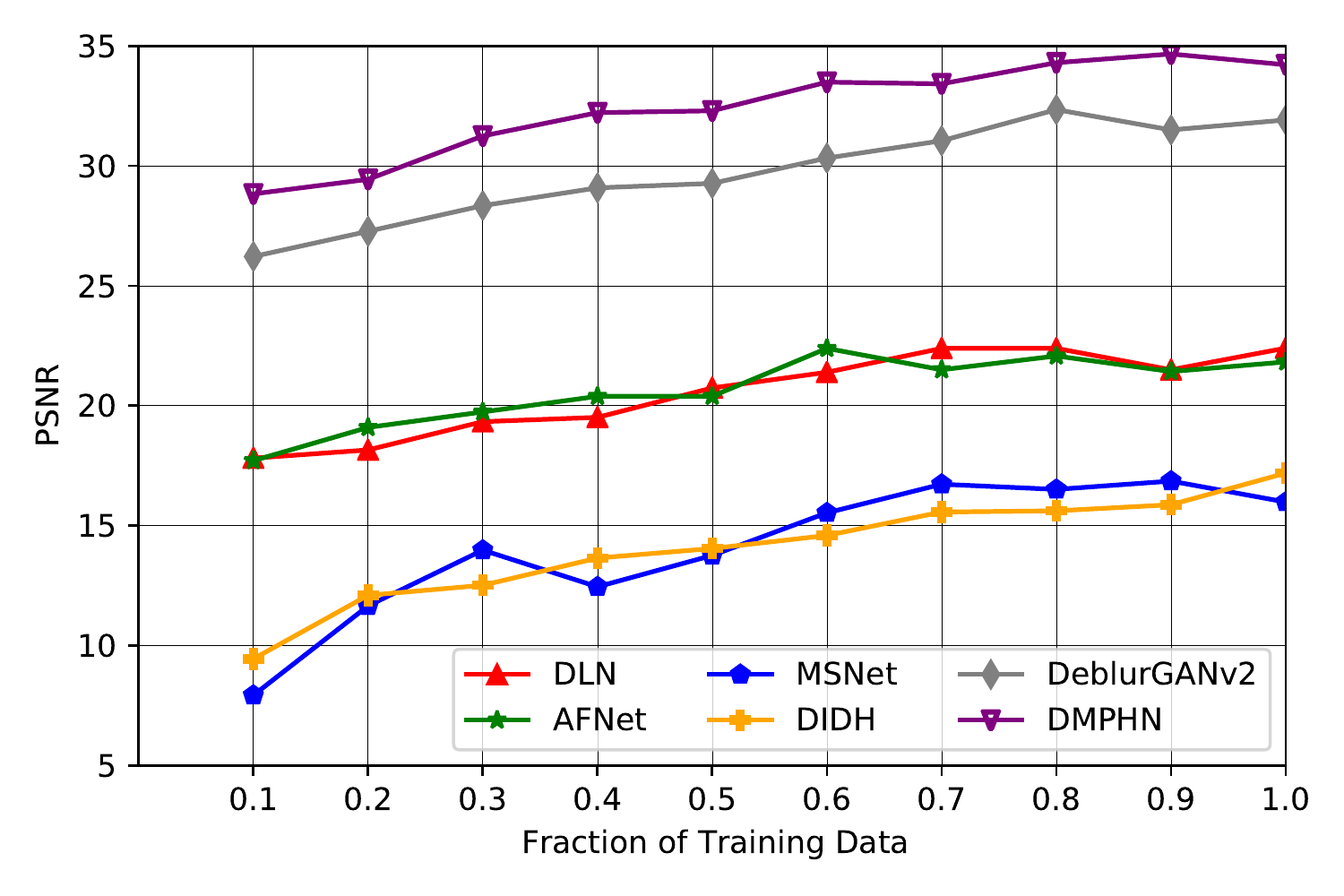} \vspace{-10pt}
\caption{Performance landscape (PSNR vs Fraction of Training Dataset) of proposed copy blend for task of LLIE, Image Dehazing and Deblurring using LOL, NTIRE-19 and GO PRO datasets respectively.}
\vspace{-2mm}
\label{fig:fig_4}
\end{figure}

\vspace{-1mm}
\subsection{Performance under reduced training data}
\vspace{-1mm}

\begin{figure}[t]
\centering
\includegraphics[width=0.99\columnwidth, height=6cm]{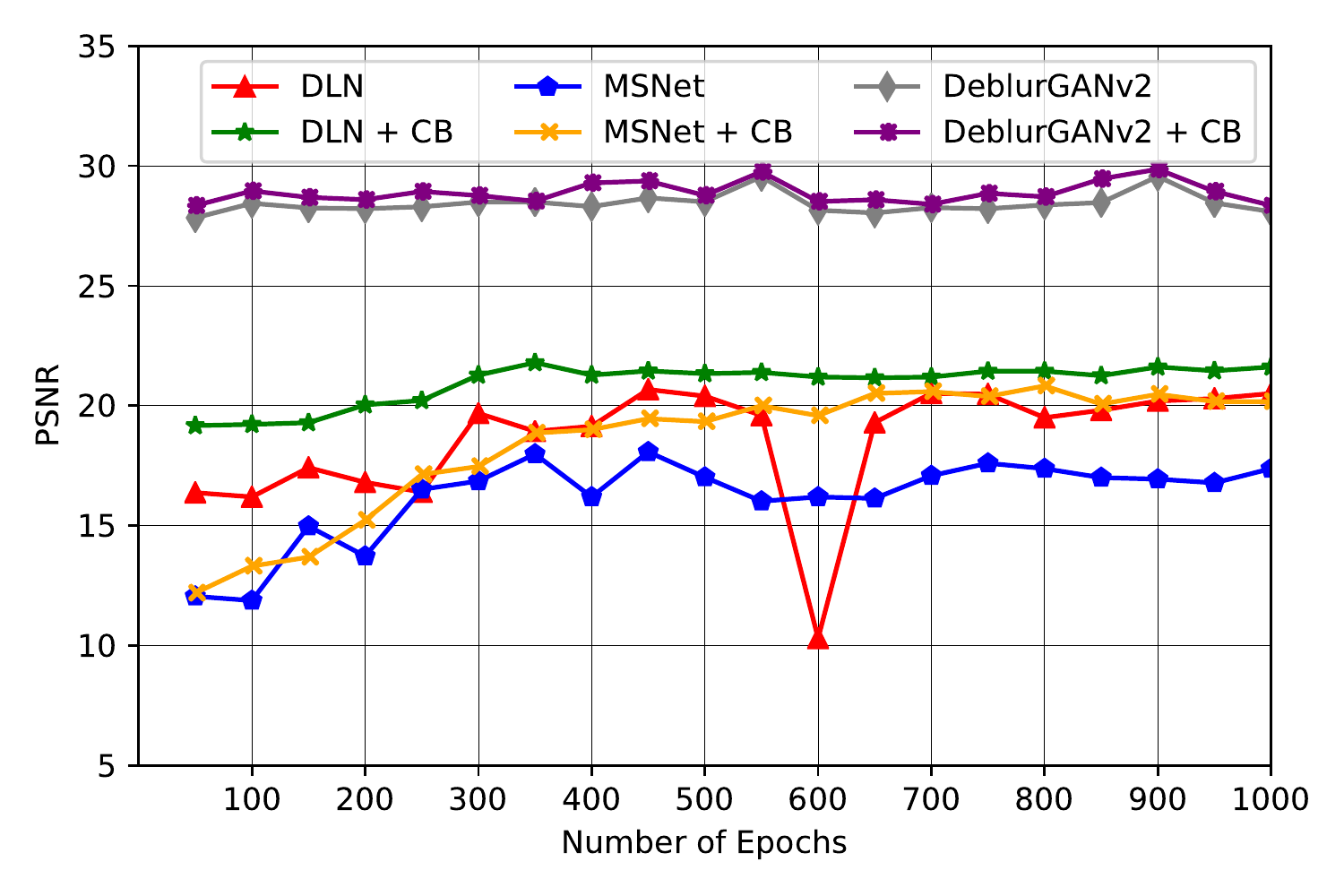} \vspace{-10pt}
\caption{Peak performance achieved with and without copy blend augmentation, when using complete dataset for training.}
\vspace{-2mm}
\label{fig:fig_5}
\end{figure}

As CNN based models are notoriously data hungry, we examine whether using augmentations such as cut-out, copy-blend and cut-mix can reduce the amount of paired training dataset while achieving peak performance. So, we retrain all algorithms from scratch and vary training dataset size from 20\% with incremental steps of 10\% of the original dataset, while keeping the number of training epochs at 1000 for LLIE, deblurring and dehazing. We observe (Fig. \ref{fig:fig_4}) these augmentation techniques to aid in achieving peak model performance across all algorithms, while relying on reduced training data, thus allowing in developing data efficient algorithms without any overfitting. Furthermore, we observe the algorithm optimization to be achieved in significantly less number of epochs as summarized in Fig. \ref{fig:fig_5}.

\vspace{-1mm}
\subsection{Varying Augmentation Shapes and Number of Patches}
\begin{figure}[t]
\centering
\includegraphics[width=0.99\columnwidth, height=6cm]{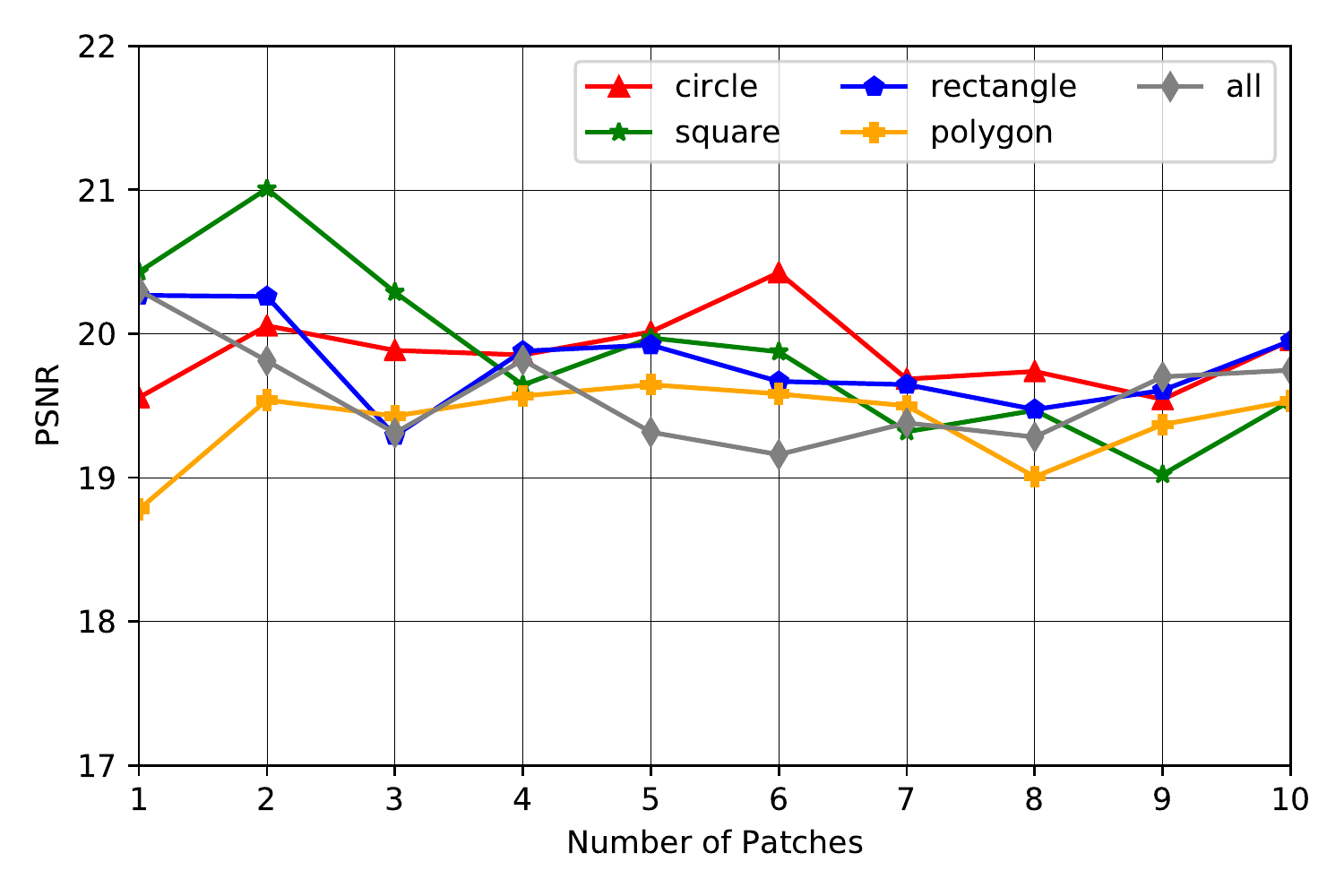} \vspace{-10pt}
\caption{Peak performance achieved by varying patch size and number of patches}
\vspace{-2mm}
\label{fig:fig_6}
\end{figure}

\vspace{-1mm}
While we fixed the number of patches and shapes, in this section we examine the effect of increasing number of patches and varying the shapes of augmentation. Based on the performance summary in Fig. \ref{fig:fig_6}, we observe the peak performance to be achieved when the number of patches is 2 and shape as square. 

\vspace{-2mm}
\section{Conclusion}
\vspace{-1mm}
In this paper we compared different region modification based augmentation approaches along with their extension, copy blend, as a data augmentation technique for low level vision tasks. We also evaluated their impact on number of images within training set, number of training epochs on model performance, without any modifications to underlying CNN.

\bibliographystyle{IEEEbib}
\bibliography{egbib}

\end{document}